\pdfoutput=1

\documentclass[11pt]{article}

\usepackage{acl2023}

\usepackage{times}
\usepackage{latexsym}
\usepackage{amsmath}
\usepackage{amssymb}
\usepackage{booktabs}
\usepackage{multirow}
\usepackage{tabularx}
\usepackage{graphicx}
\usepackage{subfigure}
\usepackage{comment}
\usepackage{ulem}

\usepackage[T1]{fontenc}

\usepackage[utf8]{inputenc}

\usepackage{microtype}

\makeatletter
\renewcommand{\@thesubfigure}{\hskip\subfiglabelskip}
\makeatother

\title{RAPS: A Novel Few-Shot Relation Extraction Pipeline with Query-Information Guided Attention and Adaptive Prototype Fusion}

\author{
Yuzhe Zhang$^{\spadesuit\diamondsuit}$, Min Cen$^{\clubsuit\diamondsuit}$\thanks{Equally Contributed.}, Tongzhou Wu$^{\clubsuit\diamondsuit}$, Hong Zhang$^{\spadesuit\star}$\thanks{Corresponding author.} \vspace{1.5mm} \\
\normalsize $^{\spadesuit}$School of Management, University of Science and Technology of China\\
\normalsize $^{\clubsuit}$School of Data Science, University of Science and Technology of China
\\
\normalsize{
\texttt{
$^\diamondsuit$\{zyz2020, cenmin0127, tzwu\}@mail.ustc.edu.cn}}\\
\normalsize{
\texttt{$^{\star}$zhangh@ustc.edu.cn}}
}

\begin{document}
\maketitle
\begin{abstract}
Few-shot relation extraction (FSRE) aims at recognizing unseen relations by learning with merely a handful of annotated instances. To generalize to new relations more effectively, this paper proposes a novel pipeline for the FSRE task based on que\textbf{R}y-information guided \textbf{A}ttention and adaptive \textbf{P}rototype fu\textbf{S}ion, namely \textbf{RAPS}. Specifically, RAPS first derives the relation prototype by the query-information guided attention module, which exploits rich interactive information between the support instances and the query instances, in order to obtain more accurate initial prototype representations. Then RAPS elaborately combines the derived initial prototype with the relation information by the adaptive prototype fusion mechanism to get the integrated prototype for both train and prediction. Experiments on the benchmark dataset FewRel 1.0 show a significant improvement of our method against state-of-the-art methods.
\end{abstract}

\section{Introduction}
\label{sec1}
Relation extraction (RE) is an important part of information extraction in the field of natural language processing \citep{bach2007review}. It aims at extracting and classifying the relation between two entities contained in a given text and can be applied in other advanced tasks \citep{li2021graph, hu2021word}, such as knowledge graph \citep{zhao-etal-2020-connecting}, machine reading comprehension \citep{ding-etal-2019-cognitive, dua-etal-2020-benefits}, and question answering \citep{karpukhin-etal-2020-dense, zhang2021bayesian}. Currently, most researches on RE start from deep learning methods, but these methods rely on large-scale and high-quality annotated datasets. In many real-world applications, it is not possible to collect sufficient instances for model training, which makes them difficult to apply. In order to solve the problem of data scarcity, Few-shot relation extraction (FSRE) task has been widely studied in recent years. In the task, a model is first trained on a large-scale annotated data with known relation types, and then quickly adapts to a small amount of data with new relation types.

Recently, many approaches have been proposed for addressing FSRE problems \citep{DBLP:conf/icml/QuGXT20, peng-etal-2020-learning, wang-etal-2020-learning-decouple, DBLP:conf/cikm/YangZDHHC20, han2021exploring}. One of the most popular algorithms is Prototypical Network \citep{DBLP:conf/nips/SnellSZ17}, which is a metric-based meta-learning framework. The main idea of the prototypical network is that each relation class has a prototype, and the prototype can be learned in the embedding space with given instances (generally average the embedding of the instances in each relation class). Finally all query instances are classified via the nearest neighbor rule.

\begin{table}[t]
	\centering
	\small
	\scalebox{0.9}{
	\begin{tabular}{p{8cm}}
		\hline
		\textbf{Support Set} \\ \hline
		\textbf{Class 1} \emph{~mother}: \\
		\qquad \textbf{Instance\;1}~Jinnah and his wife \lbrack Rattanbai Petit\rbrack$_{e_h}$ had separated soon after their daughter, \lbrack Dina Wadia\rbrack$_{e_t}$ was born. \\
		\qquad \textbf{Instance\;2}~She married (and murdered) \lbrack Polyctor\rbrack$_{e_h}$, son of Aegyptus and \lbrack Caliadne\rbrack$_{e_t}$. Apollodorus.\\
		\textbf{Class 2} \emph{~follows}: ... \\
		\textbf{Class 3} \emph{~crosses}: ... \\ \hline
		\textbf{Query Instance} \\ \hline
		Dylan and \lbrack Caitlin\rbrack$_{e_h}$ brought up their three children, \lbrack Aeronwy\rbrack$_{e_t}$, Llewellyn and Colm. \\
		\hline
	\end{tabular}}
	\caption{
		An example of 3-way 2-shot relation classification scenario from the FewRel validation set. $e_h$ marks the head entity, and $e_t$ marks the tail entity. The query instance is of Class 1: \emph{mother}. The support instances of Classes 2 and 3 are omitted.
	}
	\label{FewShotRCExample}
\end{table}

Two main lines of research tracks are adopted to improve the FSRE performance. 
The first line is to integrate the relation information (i.e., relation labels or descriptions) into the model as the external knowledge to assist prototype representation learning. 
\citet{DBLP:conf/cikm/YangZDHHC20} proposed TD-Proto model, which is an enhanced prototypical network with both relation and entity descriptions. \citet{wang-etal-2020-learning-decouple} proposed CTEG model that learns to decouple relations by adding two types of external information. 
The second line starts from the model structure or training strategy to make model learn good prototypes, that is, to learn intra-class similarity and inter-class dissimilarity. \citet{DBLP:conf/icml/QuGXT20} introduced a global relation graph into the Bayesian meta-learning framework, which makes the model better generalize across different relations. \citet{peng-etal-2020-learning} proposed a contrastive pre-training framework for RE to enhance the ability to grasp entity types and extract relational facts from contexts. \citet{han2021exploring} introduced a novel supervised contrastive learning method that obtains better prototypes by combining the prototypes, relation labels and descriptions to support model training, and designed a task adaptive focal loss to improve the performance on hard FSRE task.

However, there are two limitations in existing works. Primarily, these prototypical-network-based methods tend to construct the class prototypes simply by averaging representation of support instances of each class, which ignores the informative interaction between the support instances and the query instances. Secondly, in order to learn better representations, these works usually adopt complicated designs or networks, like graphs \citep{DBLP:conf/icml/QuGXT20}, hybrid features \citep{han2021exploring}, contrastive learning \citep{wang-etal-2020-learning-decouple, han2021exploring}, or elaborate attention networks \citep{yang-etal-2021-entity}, which may bring too many useless or even harmful parameters.  

To address aforementioned issues, this paper proposes a novel pipeline for the FSRE task based on que\textbf{R}y-information guided \textbf{A}ttention and adaptive \textbf{P}rototype fu\textbf{S}ion, namely \textbf{RAPS}. Concretely, RAPS exploits rich interactive information between the support instances and the query instances by a query-information guided attention module to obtain more accurate relation prototype representations. Furthermore, it elaborately combines the derived relation prototype with the relation information by the adaptive prototype fusion mechanism, which provides more degrees of freedom to learn from data and adjust the weights that the relation prototype and relation information hold, to get the final relation prototype.
In this way, the model gains diverse and discriminative prototype
representations without introducing too much parameters or computational-demanding modules.
Extensive experiments on FewRel \citep{han2018fewrel} benchmark show that our model significantly outperforms the baselines. Ablation and case studies demonstrate the effectiveness of the proposed modules. Our code is available at \url{https://github.com/zyz0000/RAPS}.

The contributions of this paper are summarized as follows:
\begin{itemize}
    \item We exploit the rich interactive information between the support set and the query set by the proposed query-information guided attention module to get more accurate prototype.
    \item We present a novel prototype attention fusion mechanism to further combine the useful information from the relation prototypes and the relation information.
    \item Qualitative and quantitative experiments on FewRel benchmark demonstrate the effectiveness of the proposed RAPS model.
\end{itemize}

\section{Task Definition}
\label{sec2}
We follow the typical $N$-way $K$-shot FSRE task setting, which contains a support set $\mathcal{S}$ and a query set $\mathcal{Q}$. The support set $\mathcal{S}$ includes $N$ novel classes (relations), each with $K$ labeled instances. The query set $\mathcal{Q}$ contains the same $N$ classes as $\mathcal{S}$. The task aims to predict the relation of instances in query set $\mathcal{Q}$. In addition, an auxiliary dataset $\mathcal{D}_{\text{base}}$ containing abundant base classes, each of which has a large number of labeled examples, is provided. Note that the relations in $\mathcal{D}_{\text{base}}$ and $\mathcal{S}$ are disjoint. The learner aims to acquire knowledge from the classes in $\mathcal{D}_{\text{base}}$ and make fast adaptation on novel classes in $\mathcal{S}$. Specifically, in each training iteration, $N$ different classes are randomly selected from $\mathcal{D}_{\text{base}}$ to form the support set $\mathcal{S}=\left\{s_{k}^{i}; 1 \leq i \leq N, 1 \leq k \leq K\right\}$. Meanwhile, $|\mathcal{Q}|$ instances are sampled from the remaining data of the same $N$ classes to form a query set $\mathcal{Q}=\left\{q_j; 1 \leq j \leq |\mathcal{Q}|\right\}$. Each instance in $\mathcal{D}_{\text{base}}$ can be represented as a triple $(s, e, r)$, where $s$ is a sentence of length $T$, $e= (e_1, e_2)$ is the head and tail entities and $r$ is the semantic relation between $e_1$ and $e_2$ conveyed by $s$, $r \in \mathcal{R}$, where $\mathcal{R} = \left\{r_1, \dots, r_N\right\}$ is the set of all candidate relation classes. Table \ref{FewShotRCExample} is an example of a 3-way 2-shot FSRE task.

\section{Methodology}
\label{sec3}

\begin{figure*}[htbp]
    \centering
    \includegraphics[scale=0.5]{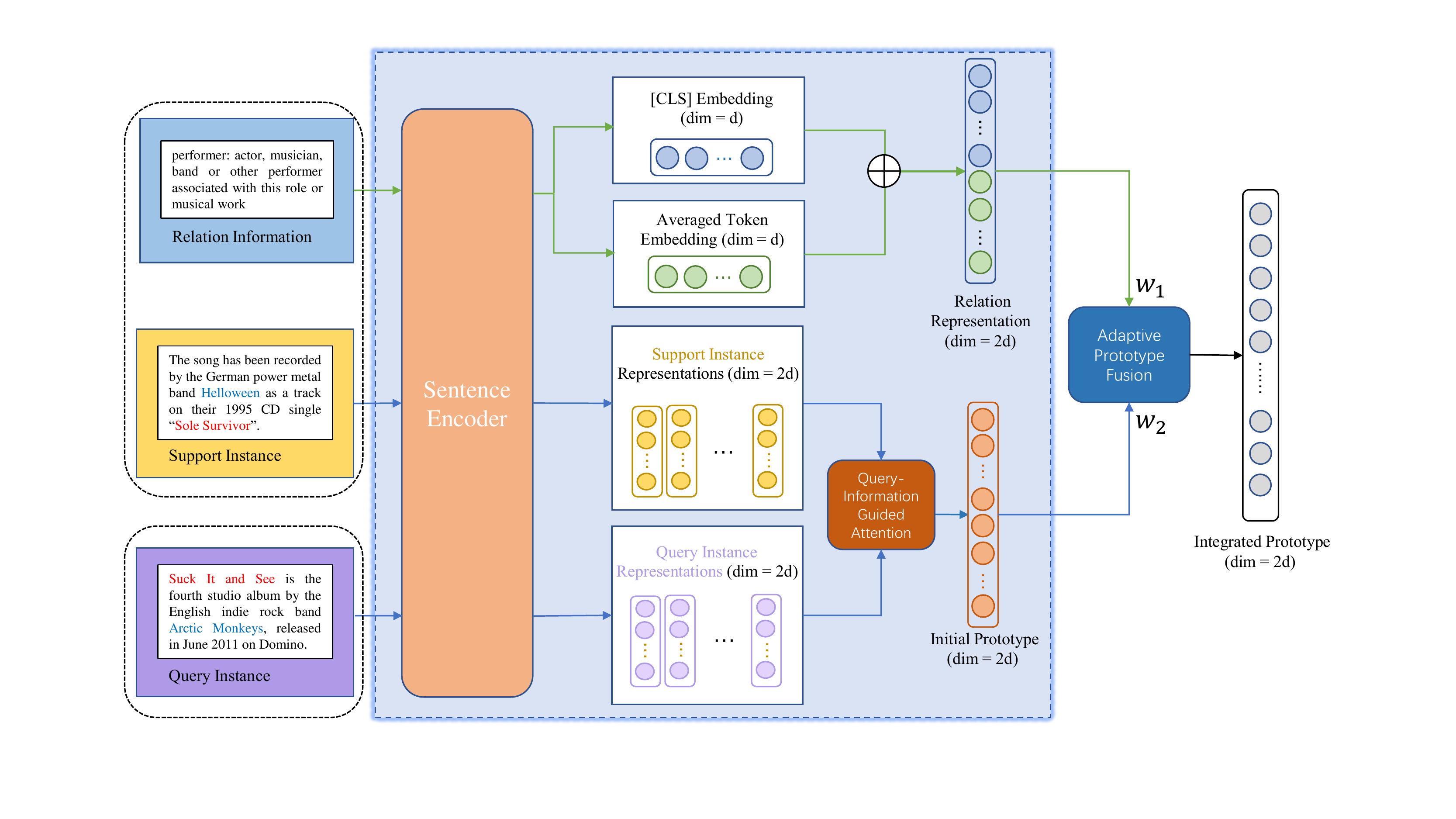}
    \caption{The pipeline of our proposed RAPS model. $\bigoplus$ represents direct concatenation.}
    \label{pipeline}
\end{figure*}

This section provides the details of our proposed RAPS. Figure \ref{pipeline} shows the overall model structure. The inputs are $N$-way $K$-shot tasks sampled from $\mathcal{D}_{\text{base}}$, where each task contains a support set and a query set. Meanwhile, we combine the names and descriptions of these $N$ relation classes as inputs as well.
We get the integrated relation prototypes with abundant information by the following four steps:
First, we encode the sentences and relation information into their embeddings by a shared encoder. 
Second, we concatenate two aspects of the relation embeddings to obtain the same dimension as instance representations.
Third, we calculate the initial prototype of each relation class by the query-information guided attention module. 
Last, we integrate relation representations into the initial prototypes by the adaptive prototype fusion mechanism to get the integrated relation prototypes.

\subsection{Sentence Encoder}
\label{sub31}
We employ BERT \citep{kenton2019bert} as the encoder to map the instances into a low-dimensional vector space and better capture the semantic information of support set $\mathcal{S}$ and $\mathcal{Q}$. For instances in $\mathcal{S}$ and $\mathcal{Q}$, we concatenate the hidden states corresponding to start tokens of two entity mentions following \citet{soares2019matching}, i.e., $h=\text{concat}(h_1, h_2) \in \mathbb{R}^{2d},$ where $h_i \in \mathbb{R}^{d}, i=1, 2$ and $d$ is the size of the representation vector (or euqally, the hidden size of BERT output), as the instance representation.

\subsection{Relation Representation}
\label{sub32}
For each relation, we combine its name and description and feed the sequence into the BERT encoder. We treat the embeddings of the [CLS] tokens $\left\{\mathbf{r}_{i}^{1}, 1 \leq i \leq N \right\}$, and the average embeddings of all tokens $\left\{\mathbf{r}_{i}^{2}, 1 \leq i \leq N \right\}$, as two components of the relation representation. Then, the final representation of each relation is derived by direct concatenation of $\mathbf{r}_{i}^{1}$ and $\mathbf{r}_{i}^{2}$, i.e., $$\mathbf{r}_{i}=\text{concat}(\mathbf{r}_{i}^{1}, \mathbf{r}_{i}^{2}) \in \mathbb{R}^{2d}.$$

\subsection{Query-Information Guided Attention Module}
\label{sub33}
Traditionally, the na\"ive prototype for relation $i$ is computed simply by averaging the representations of the $K$ instances under relation $i$ in $\mathcal{S}$ as 
\begin{equation}
    \label{average}
    \mathbf{p}_i=\frac{1}{K} \sum_{k=1}^{K} \mathbf{s}_{k}^{i}, 
\end{equation}
where $\mathbf{s}_{k}^{i}$ is the embedding of the $k$th support instance of relation $i$. However, simply averaging all support instances as the relation prototype regardless of the query instances will lose abundant interaction between them. In other words, different relation prototypes will be computed based on the semantic relevance between the support instances and the query instances.

In order to generate more informative prototypes, we propose a query-information guided attention module. This specific module aims to utilize the semantic relevance between the support instances and the query instances to help compute the customized relation prototypes. Specifically, the prototype can be represented as:
\begin{equation}
    \label{attention}
    \mathbf{p}_i=\sum_{k=1}^{K}\alpha_{k}^{i} \mathbf{s}_{k}^{i},
\end{equation}
where $\mathbf{p}_i$ is the attention-based prototype of relation $i$, $\mathbf{s}_{k}^{i}$ is the representation of the $k$th support instance of relation $i$, $\alpha_{k}^{i}$ is the weight indicating the semantic relevance between the $k$th support instance of relation $i$ and the query instances. The weight $\alpha_{k}^{i}$ is computed by
\begin{equation}
    \label{att-score}
    \alpha_{k}^{i}=\frac{\exp\left(-\frac{1}{|\mathcal{Q}|}\sum_{j=1}^{|\mathcal{Q}|} d(\mathbf{s}_{k}^{i},\mathbf{q}_{j})\right)}{\sum_{k=1}^{K}\exp\left(-\frac{1}{|\mathcal{Q}|}\sum_{j=1}^{|\mathcal{Q}|}d(\mathbf{s}_{k}^{i}, \mathbf{q}_{j})\right)},
\end{equation}
where $\mathbf{q}_j$ is the embedding of the $j$th query instance in the query set, $|\mathcal{Q}|$ is the total number of instances in $\mathcal{Q}$, and $d(\cdot, \cdot)$ means the Euclidean distance.  We call the prototype obtained in this way \textit{Initial Prototype}, which is able to store some knowledge about the query.
The benefit of the initial prototype for subsequent classification will be proved in Subsection \ref{sub44}.

\subsection{Adaptive Prototype Fusion}
\label{sub34}
Inspired by the adaptively spatial feature fusion approach proposed by \citet{liu2019asff}, we design a novel adaptive prototype fusion mechanism to obtain the \textit{integrated prototype}. Specifically, the integrated prototype $\mathbf{p}_{i}^{w}$ is an weighted average of $\mathbf{p}_i$ and $\mathbf{r}_i$: 
\begin{equation}
    \label{apf}
    \mathbf{p}_{i}^{w}=w_1 \mathbf{p}_i + w_2 \mathbf{r}_i,
\end{equation}
where $w_1, w_2 \in \mathbb{R}$ are two learnable weights, which depend on the representation of category information to ensure that the calculated integrated prototype $\mathbf{p}_{i}^{w}$ is compatible and not redundant. 
Unlike \citet{liu2019asff} forces the sum of the weights equal to 1, our adaptive prototype fusion has no constraint on $w_1$ and $w_2$. This provides more degrees of freedom to learn from data and adjust the weights. The benefits of mechanism will be demonstrated in Subsection \ref{sub45}.

\subsection{Training Objective}
\label{sub35}
With the representation of query and integrated prototypes of $N$ relations, the model uses the vector dot product way to compute the probability of the relations for the query instance representation $\mathbf{q}_j$ as follows:
$$P(y=i | \mathbf{q}_j) = \frac{\exp(\mathbf{q}_{j} \cdot \mathbf{p}_{i}^{w})}{\sum_{n=1}^{N} \exp(\mathbf{q}_{j} \cdot \mathbf{p}_{n}^{w})}.$$
The training loss $\mathcal{L}$ is defined as regular cross entropy loss as follows:
$$\mathcal{L}=-\sum_{j=1}^{|\mathcal{Q}|} \log \frac{\exp(\mathbf{q}_{j} \cdot \mathbf{p}_{i}^{w})}{\sum_{n=1}^{N} \exp(\mathbf{q}_{j} \cdot \mathbf{p}_{n}^{w})}.$$
In the prediction stage, query $\mathbf{q}_j$ is assigned to relation $i$ with the highest probability:
$$\text{label}=\arg \max_{i} P(y=i | \mathbf{q}_j).$$

\section{Experiments}
\label{sec4}
\subsection{Datasets and Baselines}
\label{sub41}

\textbf{Datasets} 
We use FewRel 1.0 \citep{han2018fewrel} and FewRel 2.0 (the domain adaption portion) \citep{gao-etal-2019-fewrel} to evaluate our model. FewRel 1.0 is a large-scale FSRE dataset, which contains 100 relations with 700 instances each relation. We follow the official split to use 64, 16 and 20 relations for training, validation, and testing. In order to study the domain transferability of RAPS, we also evaluate our model on FewRel 2.0. The training set of FewRel 2.0 is the same as FewRel 1.0, and the validation set is collected from the biomedical domain which contains 10 relations and 100 instances for each relation. We evaluate our model on FewRel 1.0 and FewRel 2.0 in terms of the accuracy under multiple $N$-way $K$-shot meta tasks. We select $N$ to be 5 and 10, $K$ to be 1 and 5 to form four test scenarios according to \citet{DBLP:conf/aaai/GaoH0S19}.

\begin{table*}[htbp]
	\centering
	\scalebox{0.9}{
		\begin{tabular}{clcccc}
			\toprule
			Encoder & Model&5-way-1-shot&5-way-5-shot&10-way-1-shot&10-way-5-shot\\
			\midrule
			\multirow{3}*{\rotatebox{90}{CNN}} &Proto-CNN$^\clubsuit$ \citep{DBLP:conf/nips/SnellSZ17} &72.65 / 74.52 &86.15 / 88.40 &60.13 / 62.38 &76.20 / 80.45 \\
			&Proto-HATT \citep{DBLP:conf/aaai/GaoH0S19} &75.01 / --- --- & 87.09 / 90.12 &62.48 / --- ---&77.50 / 83.05 \\
			&MLMAN \cite{ye-ling-2019-multi} &79.01 / 82.98 & 88.86 / 92.66 &67.37 / 75.59&80.07 / 87.29 \\
			\midrule
			\multirow{14}*{\rotatebox{90}{BERT}} &Proto-BERT$^\ast$ \citep{DBLP:conf/nips/SnellSZ17} &82.92 / 80.68 &91.32 / 89.60 &73.24 / 71.48 &83.68 / 82.89 \\
			&MAML$^\ast$ \cite{DBLP:conf/icml/FinnAL17} &82.93 / 89.70 & 86.21 / 93.55 &73.20 / 83.17&76.06 / 88.51 \\
			&GNN$^\ast$ \cite{DBLP:conf/iclr/SatorrasE18} &--- --- / 75.66 & --- --- / 89.06 &--- --- / 70.08&--- --- / 76.93 \\
			&BERT-PAIR$^\clubsuit$ \cite{gao-etal-2019-fewrel} &85.66 / 88.32 & 89.48 / 93.22 &76.84 / 80.63&81.76 / 87.02 \\
			&REGRAB \cite{DBLP:conf/icml/QuGXT20} & 87.95 / 90.30 & 92.54 / 94.25 &80.26 / 84.09&86.72 / 89.93 \\
			&TD-Proto \cite{DBLP:conf/cikm/YangZDHHC20} & --- --- / 84.76 &  --- --- / 92.38 &--- --- / 74.32& --- --- / 85.92 \\
			&CTEG \cite{wang-etal-2020-learning-decouple} &84.72 / 88.11 & 92.52 / 95.25 &76.01 / 81.29&84.89 / 91.33 \\
			&ConceptFERE \cite{yang-etal-2021-entity} &--- --- / 89.21 &--- --- / 90.34 &--- --- / 75.72 &--- --- / 81.82 \\
			&HCRP \cite{han2021exploring} & 90.90 / 93.76   & 93.22 / 95.66   & 84.11 / 89.95 & 87.79 / 92.10 \\
			&DRK \cite{wang2022drk}  & --- --- / 89.94 & --- --- / 92.42 & --- --- / 81.94 & --- --- / 85.23  \\ 
			&\textbf{RAPS} & \textbf{92.26 / 94.93}  & \textbf{94.08 / 96.92} & \textbf{87.23 / 90.65} & \textbf{89.87 / 93.72} 
			\\ \cline{2-6}
			&MTB \cite{soares2019matching}  & --- --- / 91.10  & --- --- / 95.40  & --- --- / 84.30 & --- --- / 91.80 \\
            &CP \cite{peng-etal-2020-learning}  & --- --- / 95.10  & --- --- / 97.10 & --- --- / 91.20  & --- --- / 94.70 \\
            &MapRE \cite{dong-etal-2021-mapre} & --- --- / 95.73 & --- --- / 97.84 & --- --- / 93.18  & --- --- / 95.64 \\
            & HCRP (CP) & 94.10 / 96.42 & 96.05 / 97.96 & 89.13 / 93.97 & 93.10 / \textbf{96.46} \\
            & \textbf{RAPS (CP)} & \textbf{96.28 / 97.39} &  \textbf{97.74 / 98.00} & \textbf{93.86 / 95.21} & \textbf{95.39} / 96.32  \\
			\bottomrule
		\end{tabular}
	}
	\caption{Accuracy (\%) of few-shot classification on the FewRel 1.0 validation / test set. $^\clubsuit$The results of Proto-CNN and BERT-PAIR are from FewRel public leaderboard (\url{https://thunlp.github.io/fewrel.html}), $^\ast$the results of Proto-BERT and MAML are reported in \citet{DBLP:conf/icml/QuGXT20}.}
	\label{main-fewrel1}
\end{table*}

\noindent \textbf{Baselines} We compare our model with the following baseline
methods:
1) \textbf{Proto-CNN} \citep{DBLP:conf/nips/SnellSZ17}, prototypical networks with CNN encoder. 
2) \textbf{Proto-HATT} \citep{DBLP:conf/aaai/GaoH0S19}, hybrid attention is applied on prototypical networks to focus on the crucial instances and features. 
3) \textbf{MLMAN} \citep{ye-ling-2019-multi}, a multi-level matching and aggregation prototypical network. 
4) \textbf{Proto-BERT} \citep{DBLP:conf/nips/SnellSZ17}, prototypical networks with BERT encoder. 
5) \textbf{MAML} \citep{DBLP:conf/icml/FinnAL17}, the model-agnostic meta-learning algorithm. 
6) \textbf{GNN} \citep{DBLP:conf/iclr/SatorrasE18}, a meta-learning approach using graph neural networks. 
7) \textbf{BERT-PAIR} \citep{gao-etal-2019-fewrel}, a method that measures the similarity of sentence pairs. 
8) \textbf{REGRAB} \citep{DBLP:conf/icml/QuGXT20}, a Bayesian meta learning method with an external global relation graph. 
9) \textbf{TD-Proto} \citep{DBLP:conf/cikm/YangZDHHC20}, learning the importance distribution of generic content words by a memory network.
10) \textbf{CTEG} \citep{wang-etal-2020-learning-decouple}, a model using dependency trees to learn to decouple high co-occurrence relations, where two external information are added. 
11) \textbf{ConceptFERE} \citep{yang-etal-2021-entity}, introducing the inherent concepts of entities to provide cludes for relation prediction. 
12) \textbf{HCRP} \citep{han2021exploring}, introducing Hybrid Prototype Learning, Relation-Prototype Contrastive Learining, and Taks Adaptive Focal Loss for the model improvement.
13) \textbf{DRK} \citep{wang2022drk}, introducing a logic rule to constrain the inference process, thereby avoiding the adverse effect of shallow text features.
Moreover, we compare our model with three pretrained RE methods: 
13) \textbf{MTB} \citep{soares2019matching}, pretrained by matching the blank stategy on top of an existing BERT model. 
14) \textbf{CP} \citep{peng-etal-2020-learning}, an entity masked contrastive pretraining framework for RE while utilizing prototypical networks for finetuning on FSRE. 
15) \textbf{MapRE} \citep{dong-etal-2021-mapre}, a framework considering both label-agnostic and label-aware semantic mapping information in pre-training and fine-tuning. 
Note that MTB \citep{soares2019matching} employs BERT$_{\small \texttt{LARGE}}$ as the backbone, and CP \citep{peng-etal-2020-learning} and MapRE \citep{dong-etal-2021-mapre} all employ additional pre-training on BERT with Wikipedia data or contrastive learning to get better contextual representation. This is why we do not compare with MTB, CP, and MapRE directly.

\subsection{Training and Evaluation}
\label{sub42}
\textbf{Training} We use BERT$_{\small \texttt{BASE}}$ and CP \citep{wang-etal-2020-learning-decouple} as the sentence encoder, and set the total train iteration number as 30,000, validation iteration number as 1,000, batch size as 4, learning rate as $1 \times 10^{-5}$ and $5 \times 10^{-6}$ for BERT and CP respectively.

\noindent \textbf{Evaluation} For FewRel 1.0, we report the accuracy on validation and test sets. For FewRel 2.0, we report the accuracy on test set. Since the labels of two test sets are not publicly available, we submit the prediction file of our best model to the CodaLab platform \footnote{\url{https://codalab.lisn.upsaclay.fr}} to obtain the final result on the test set.

\subsection{Overall Evaluation Results}
\label{sub43}
Table \ref{main-fewrel1} presents the experimental results on FewRel 1.0 validation set and test set. As shown in the upper part of Table \ref{main-fewrel1}, our method outperforms the strong baseline models significantly. To be specific, RAPS achieves the average of 1.19 points improvement in terms of accuracy on the test sets of four meta tasks, compared to the second best method (HCRP), demonstrating the superior generalization ability.
In addition, we evaluate our approach based on the model CP \citep{peng-etal-2020-learning}, where the BERT encoder is initialized with the pre-trained parameters by a contrastive pre-training approach. The lower part of Table \ref{main-fewrel1} shows that our approach achieves a consistent performance boost when using CP pre-trained model. It is worth to mention that RAPS (CP) has 1.17 and 1.24 points improvement compared with HCRP (CP), under the 5-way 1-shot and 10-way 1-shot scenarios, respectively, which demonstrates that our approach is more suitable for few-shot scenarios. 
Our method also achieves the competitive performance compared with HCRP on FewRel 2.0, as shown in Table \ref{main-fewrel2}, which shows that RAPS is also transferable to the domain adaptation setting. It can be seen that RAPS is overall better than HCRP with the CP as encoder. 
The possible reason why RAPS is worse than HCRP when using BERT as encoder is that the FewRel 2.0 with domain adaptation setting only provides the name of relations without a specific description, which has a relatively harmful impact on the adaptive fusion mechanism to generate a strong relation representation for the relation prototypes. Nevertheless, HCRP seems to be more robust to the lack of relation description due to the complex hybrid prototype learning module.
	
\begin{table}
	\centering
	\renewcommand\tabcolsep{3.6pt}
	\scalebox{0.8}{
		\begin{tabular}{lcccc}
			\toprule
			\multirow{2}*{Model} 
			&5-way&5-way&10-way& 10-way\\
			&1-shot&5-shot&1-shot& 5-shot\\
			\midrule 
			Proto-CNN$^\ast$ &35.09 &49.37  &22.98  &35.22  \\
			Proto-BERT$^\ast$   &40.12  & 51.50  &26.45 &36.93  \\
			BERT-PAIR$^\ast$ & 67.41 & 78.57 & 54.89 & 66.85 \\
			Proto-CNN-ADV$^\ast$  & 42.21 &  58.71 &28.91 & 44.35 \\
			Proto-BERT-ADV$^\ast$  & 41.90 &  54.74 &27.36 & 37.40 \\
			DaFeC+BERT-PAIR$^\clubsuit$ & 61.20 & 76.99 & 47.63 & 64.79 \\
			Cluster-ccnet$^\dagger$ & 67.70 & 84.30 & 52.90 & 74.10 \\
			HCRP$\spadesuit$ & \textbf{76.34} & 83.03 & \textbf{63.77} & 72.94 \\
			\hline
			RAPS & 74.99 & \textbf{87.85} & 60.29 & \textbf{80.10} \\
			\textbf{RAPS} (CP) & \textbf{80.61} & \textbf{89.59} & \textbf{67.51} & \textbf{82.52} \\
			\bottomrule
		\end{tabular}
	}
	\caption{Accuracy (\%) of few-shot classification on the FewRel 2.0 domain adaptation test set. $^\ast$The results are quoted from FewRel leaderboard\footnote{\url{https://thunlp.github.io/fewrel.html}}, $^\clubsuit$the results are quoted from \citet{cong2020inductive}, $^\spadesuit$the results are quoted from \citet{han2021exploring}, $^\dagger$the results are quoted from \citet{bansal-etal-2021-diverse}.}
	\label{main-fewrel2}
\end{table}

\subsection{Effects of Different Adaptive Prototype Fusion Methods}
\label{sub44}
In this subsection, we further explore four different types of adaptive prototype fusion methods, including:
\begin{itemize}
    \item Unconstrained Adaptive Scalar (UAS): $\mathbf{p}_{i}^{w}=w_1 \mathbf{p}_i + w_2 \mathbf{r}_i, w_1, w_2 \in \mathbb{R}$.
    \item Constrained Adaptive Scalar (CAS): $\mathbf{p}_{i}^{w}=w_1 \mathbf{p}_i + w_2 \mathbf{r}_i, w_1, w_2 \in \mathbb{R}, \text{subject to} \, w_1 + w_2=1$.
    \item Unconstrained Adaptive Matrix (UAM): $\mathbf{p}_{i}^{W}=\mathbf{W}_1 \mathbf{p}_i + \mathbf{W}_2 \mathbf{r}_i, \mathbf{W}_1, \mathbf{W}_2 \in \mathbb{R}^{2d \times 2d}$.
    \item Constrained Adaptive Matrix (CAM): $\mathbf{p}_{i}^{W}=\mathbf{W}_1 \mathbf{p}_i + \mathbf{W}_2 \mathbf{r}_i, \mathbf{W}_1, \mathbf{W}_2 \in \mathbb{R}^{2d \times 2d}, \text{subject to} \, \mathbf{W}_1 + \mathbf{W}_2=\mathbf{I}_{2d}$.
\end{itemize}
For UAS and CAS, $w_1$ and $w_2$ are two learnable scalars. For UAM and CAM, $\mathbf{W}_1$ and $\mathbf{W}_2$ are two learnable matrices, $\mathbf{I}_{2d}$ is the identity matrix of order $2d$. Note that UAM and CAM are equivalent to add a fully-connected layer without bias term on $\mathbf{p}_i$ and $\mathbf{r}_i$, respectively. 
We evaluate these four types of adaptive prototype fusion methods on FewRel 1.0 validation set. The results are shown in Table \ref{resultapf}. We can see that UAS gets the best performance in all scenarios. Compared with CAS, UAS has more degrees of freedom to learn the trade-off between relation prototype and relation information, thus leading to more accurate prototype and better performance than CAS. It can be seen that matrix-based fusion methods (UAM and CAM) are consistently inferior to scalar-based fusion methods (UAS and CAS), which may attribute to the introduction of useless parameters ($O(d^2)$ for UAM and CAM vs. $O(1)$ for UAS and CAS). 

\begin{table}[htbp]
	\centering
	\scalebox{0.8}{
		\begin{tabular}{c|ccccc}
		\hline
		\multirow{2}*{Model} &
		\multirow{2}*{Mean}
		&5-way&5-way&10-way& 10-way \\
		& &1-shot&5-shot&1-shot& 5-shot  \\
		\hline
		UAS & 90.86 & 92.26 & 94.08 & 87.23 & 89.87 \\
		CAS & 89.79  & 91.91 & 93.09 & 86.55 & 87.61 \\
		UAM & 75.02 & 78.56 & 85.84 & 64.58 & 71.11 \\
		CAM & 73.58  & 74.64 & 84.69 & 61.37 & 73.63 \\
		\hline
	    \end{tabular}}
	\caption{Results of different prototype fusion methods under different scenarios on the FewRel 1.0 validation set (\%).}
	\vspace{-0.7cm}
	\label{resultapf}
\end{table}

\subsection{Ablation Studies}
\label{sub45}
In this subsection, we conduct an ablation study on 5-way 1-shot and 10-way-1-shot based on BERT with the validation set, to demonstrate the effectiveness of the proposed query-information guided attention module and adaptive prototype fusion mechanism (abbreviated as QIA and APF for further reference, respectively). We consider three ablation experiments including \textbf{w/o QIA}, \textbf{w/o APF}, and \textbf{w/o QIA and APF}.
\begin{itemize}
    \item w/o QIA: calculate the prototype using \eqref{average} instead of \eqref{attention} + \eqref{att-score}.
    \item w/o APF: calculate $\mathbf{p}_i^{w}$ using direct addition $\mathbf{p}_{i}^{w}=\mathbf{p}_i + \mathbf{r}_i$ instead of \eqref{apf}.
    \item w/o QIA and APF: calculate the prototype using \eqref{average} instead of \eqref{attention} + \eqref{att-score}, and calculate $\mathbf{p}_i^{w}$ using direct addition $\mathbf{p}_{i}^{w}=\mathbf{p}_i + \mathbf{r}_i$ instead of \eqref{apf}.
\end{itemize}

From the results in Table \ref{ablation}, We can obtain several observations. First, the model performance drops more or less without any of QIA or APF, which demonstrates the effectiveness of RAPS. Second, QIA seems to be more important under 10-way 1-shot scenario than 5-way 1-shot, with the loss of performance 0.15\% and 0.76\%, respectively. Third, APF seems to be more important under 5-way 1-shot scenario than 10-way 1-shot, with the loss of performance 1.35\% and 1.13\%, respectively. Based on the above discussion, it may be a future research direction to choose appropriate prototype computation method under different $N$-way $K$-shot settings.

\begin{table}[htbp]
	\centering
		\begin{tabular}{c|cc}
		\hline
		\multirow{2}*{Method} 
		&5-way&10-way\\
		&1-shot&1-shot\\
		\hline
		QIA+APF & 92.26 & 87.23  \\
		w/o QIA & 92.11  & 86.47  \\
		w/o APF & 91.91 &  86.10  \\
		w/o QIA and APF &  91.33 & 86.02  \\
		\hline
	    \end{tabular}
	\caption{Results of ablation study on FewRel 1.0 validation set (\%). w/o is the abbreviation of without.}
	\vspace{-0.7cm}
	\label{ablation}
\end{table}

\subsection{Visualization}
\label{sub46}
\begin{figure}[htbp]
    \centering
    \subfigure[]{
        \includegraphics[width=3.6cm]{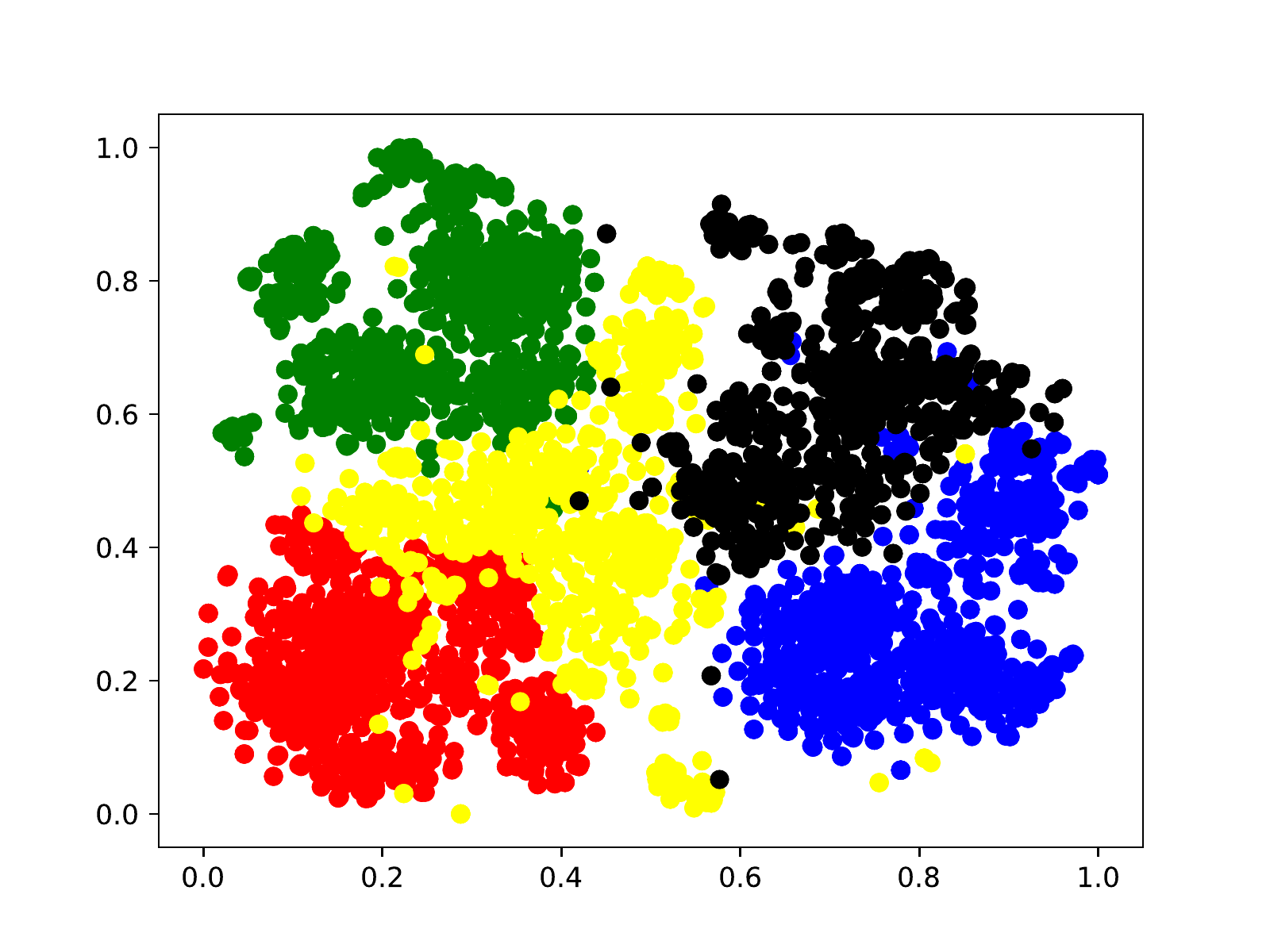}
        \label{}
        \hspace{-5mm}
    }
    \subfigure[]{
        \includegraphics[width=3.6cm]{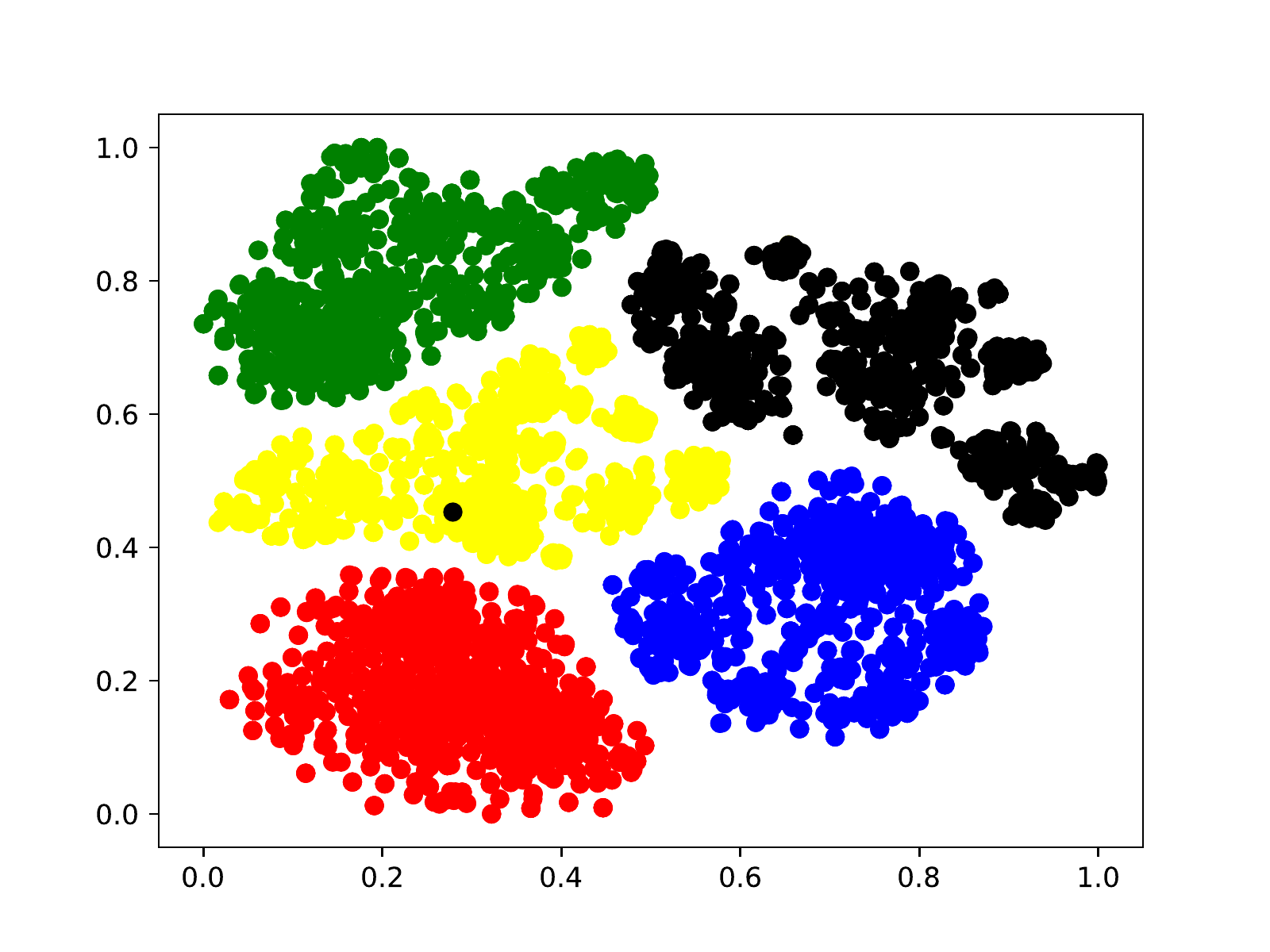}
        \label{}
    }
    \caption{Instances Visualization. Left: Instances in Proto-BERT; Right: Instances in proposed model.}
    \label{tsne-plots}
\end{figure}

In order to further explore the reason why our proposed method has excellent discriminative capability, we give the visualization results in Figure \ref{tsne-plots} with BERT on 5-way 1-shot of the validation set of FewRel 1.0. Figure \ref{tsne-plots} shows the t-SNE \citep{tsne-jmlr} visualization of query instances, where different colors represent different relation classes. The left subfigure means the original statements of query instances encoded by Proto-BERT \citep{DBLP:conf/nips/SnellSZ17}, and the right subfigure means the statements of query instances encoded by RAPS.

It can be seen from the left subfigure that, although the instances can also be divided into different classes, the intra-class distances are not close enough, and there are multiple error points (i.e., yellow points in red cluster). After introducing the relation information and training with the proposed query-information guided attention and adaptive prototype fusion mechanism (right), we can see that the error points are reduced while the representations of the same class are closer, and the inter-class boundaries are more distinct. The observation shows that our proposed method is indeed beneficial to the improvement of the model.

\section{Conclusion}
\label{sec5}
In this paper, we propose a novel FSRE pipeline using the proposed query-information guided attention module and adaptive prototype fusion mechanism, called RAPS. It has two advantages: 1) RAPS exploits rich interactive information between the support instances and the query instances to obtain more accurate initial relation prototypes. 2) RAPS dynamically makes the trade-off between the derived relation prototype and the relation information by the adaptive prototype fusion mechanism to get more compatible final relation prototype. Experiments on the FewRel 1.0 and FewRel 2.0 datasets show that RAPS achieves a significant improvement against the modern state-of-the art methods. One possible direction of future work is to generalize the adaptive prototype fusion mechanism to other text classification tasks such as intent classification.

\section*{Limitations}
There are several limitations of this work. First, RAPS only works under the $N$-way $K$-shot setup, because it requires a support set to calculate the relation class prototype. Second, its effectiveness is only examined on the task of few-shot relation extraction, while the generalization to other text classification tasks, such as intent classification and news classification, is not yet explored in this paper. Third, the model has not been extended to perform non-of-the-above (NOTA) detection \citep{gao-etal-2019-fewrel}, where a query instance may not belong to any class in the support set.

\section*{Acknowledgements}
The research of Yuzhe Zhang, Min Cen, and Tongzhou Wu is partly supported by the postgraduate studentship of USTC. Hong Zhang's research is partly supported by the National Natural Science Foundation of China (7209121, 12171451).

\bibliography{acl2023}
\bibliographystyle{acl_natbib}

\end{document}